\documentclass[11pt]{amsart}
\usepackage[a4paper, margin=2.5cm]{geometry}
\usepackage{amsmath, amssymb, amsthm, mathrsfs}
\usepackage{graphicx}
\usepackage{hyperref}
\usepackage{enumitem}
\usepackage{tikz}
\usetikzlibrary{arrows.meta, decorations.markings, calc}
\usepackage[linesnumbered,ruled,vlined]{algorithm2e}

\title{Contextuality, Holonomy and Discrete Fiber Bundles in Group-Valued Boltzmann Machines}

\author{Jean-Pierre Magnot}
\address{{LAREMA, Universit\'e d’Angers, 2 Bd Lavoisier, 
49045 Angers cedex 1, France;  Lyc\'ee Jeanne d'Arc, 40 avenue de Grande Bretagne, 63000 Clermont-Ferrand, 
France}; Lepage Research Institute, 17 novembra 1, 081 16 Presov, Slovakia}
\email{\small magnot@math.cnrs.fr; jean-pierr.magnot@ac-clermont.fr}

\begin{document}

\maketitle

\begin{abstract}
We propose a geometric extension of restricted Boltzmann machines (RBMs) by allowing weights to take values in abstract groups such as \( \mathrm{GL}_n(\mathbb{R}) \), \( \mathrm{SU}(2) \), or even infinite-dimensional operator groups. This generalization enables the modeling of complex relational structures, including projective transformations, spinor dynamics, and functional symmetries, with direct applications to vision, language, and quantum learning. 

A central contribution of this work is the introduction of a \emph{contextuality index} based on group-valued holonomies computed along cycles in the RBM graph. This index quantifies the global inconsistency or "curvature" induced by local weights, generalizing classical notions of coherence, consistency, and geometric flatness. We establish links with sheaf-theoretic contextuality, gauge theory, and noncommutative geometry, and provide numerical and diagrammatic examples in both finite and infinite dimensions.

This framework opens novel directions in AI, from curvature-aware learning architectures to topological regularization in uncertain or adversarial environments.
\end{abstract}



\section{Introduction}
The notion of \emph{contextuality} — where the outcome of a local judgement depends on the broader context in which it is evaluated — is foundational in areas ranging from quantum mechanics and logic, to decision theory and machine learning \cite{abramsky2011}.  Recently, the geometrization of such phenomena via structures on pairwise comparison matrices with entries in abstract groups has revealed deep connections between inconsistency, holonomy, and global coherence \cite{magnot2019}. In particular, the work reported at the \emph{Geometric Methods in Physics XL} workshop emphasized how problems of coherence, consistency, and uncertainty share common mathematical underpinnings across disciplines as diverse as physics, decision theory, and approximate reasoning \cite{magnot2024blind}. In parallel, efforts to model \emph{random pairwise comparisons} have highlighted the need for extending deterministic consistency measures to stochastic settings, where inconsistency becomes a distributional phenomenon with fiber-bundle-like decompositions \cite{magnot2024random}. These insights pave the way for framing contextuality as a geometric obstruction, opening new avenues in machine learning architectures that account for both noise and underlying topology. 

Standard RBMs use scalar weights and model a global probability distribution over visible and hidden binary units. In this work, we build a bridge between contextuality theory and the geometry of restricted Boltzmann machines (RBMs), inspired by earlier studies of pairwise comparisons with coefficients in a group \( G \) \cite{magnot2019}. We propose a novel class of \emph{geometric Boltzmann machines} in which weights take values in a group \( G \). In other words, we extend the RMBs formalism by allowing weights to take values in a non-necessarily commutative group \( G \), thus enriching the network with geometric and algebraic structure. The combinatorial structure of the associated bipartite graph gives rise to a discrete \( G \)-principal bundle, where holonomies along cycles encode deviations from local consistency. Thus, contextuality is directly linked to the non-triviality of this bundle, offering a coherent geometric measure suited to AI models. This enables us to model not only the intensity of interaction between units but also the structural dependencies that emerge through composition of group-valued weights.

A key object in this setting is the holonomy of a cycle---the group-valued product of weights around a loop. Holonomy captures the failure of local consistency and provides a natural measure of contextuality. By averaging holonomies over cycles, we define a global contextuality index \( \kappa \), which quantifies how far the network is from admitting a globally coherent interpretation. This parallels ideas in sheaf-theoretic contextuality, where obstructions to global sections encode logical inconsistencies \cite{abramsky2011}.

Concretely, we show:
\begin{itemize}
  \item How contextuality in RBMs can be quantified using holonomies and compactified into a scalar \emph{contextuality index} \( \kappa \), suitable for optimization.
  \item How classical notions such as Berry phases, gauge invariance, and cohomological obstructions find discrete analogues in our framework.
  \item Applications in cognitive modeling, vision, quantum machine learning, and projective inference, where geometric consistency plays a pivotal role.
\end{itemize}

By merging ideas from discrete geometry, operator algebras, and probabilistic learning, our approach provides a principled path toward interpretable and curvature-aware neural architectures — a direction ripe for exploration in the AI community.

\section{Group-Valued Boltzmann Machines as Discrete Fiber Bundles}
We consider a bipartite graph \( \mathcal{G} = (V, H, E) \), with visible units \( V \), hidden units \( H \), and weighted edges \( E \subseteq V \times H \). In standard RBMs, each edge \( (i,j) \in E \) carries a real-valued weight \( w_{ij} \in \mathbb{R} \). Here, we generalize to weights \( w_{ij} \in G \), where \( G \) is a (possibly non-abelian) group.

This construction naturally defines a discrete principal \( G \)-bundle over the vertex set \( V \cup H \), with the weights serving as local transition functions (i.e., connection data). A configuration of visible and hidden units induces a configuration of fibers, and transitions between units correspond to left multiplication by group elements.

A network is said to be \emph{coherent} if there exists a global section \( s: V \cup H \to G \) such that for all \( (i,j) \in E \), we have:
\[
    w_{ij} = s(j) s(i)^{-1}.
\]
In this case, the RBM is geometrically trivial and globally consistent.

Cycles in the graph correspond to closed paths, and their composite weights yield holonomies, which capture the failure of global triviality. This parallels the role of curvature in differential geometry and the notion of topological charge in lattice gauge theory \cite{wilson1974confinement}.

\section{Holonomy and Contextuality Index}
Let \( C = (i_1, i_2, \ldots, i_k, i_1) \) be a closed cycle in the RBM graph. We define the holonomy of \( C \) as the ordered product:
\[
    \mathrm{Hol}(C) = w_{i_1 i_2} w_{i_2 i_3} \cdots w_{i_k i_1} \in G.
\]
The cycle is said to be \emph{flat} or \emph{non-contextual} if \( \mathrm{Hol}(C) = e_G \), the identity in \( G \).

To quantify the degree of contextuality, we define a function:
\[
    \iota(C) := d_G(\mathrm{Hol}(C), e_G),
\]
where \( d_G \) is a distance function on \( G \) (e.g., the geodesic distance on a Lie group, the Cayley metric on a finite group, or the norm of the logarithm in a matrix group).

The \emph{contextuality index} of the RBM is defined by:
\[
    \kappa := \frac{1}{|\mathcal{C}|} \sum_{C \in \mathcal{C}} \iota(C),
\]
where \( \mathcal{C} \) is a family of cycles (e.g., all triangles or all simple loops up to a given length).

\subsection*{Diagrammatic Representation of Geometric RBMs}

A group-valued RBM can be represented as a discrete fiber bundle over a bipartite graph. Each unit corresponds to a node, and each weight \( w_{ij} \in G \) to an edge labeled with a group element (a connection).

The diagram has:
\begin{itemize}
    \item \textbf{Nodes}: visible (bottom) and hidden (top) units,
    \item \textbf{Edges}: labeled with group elements,
    \item \textbf{Fibers}: local group copies at each node (represented vertically),
    \item \textbf{Trivializations}: local sections attempting to assign coherent group elements to each node,
    \item \textbf{Holonomies}: closed loops illustrating the failure of triviality.
\end{itemize}

This interpretation mirrors:
\begin{itemize}
    \item parallel transport in gauge theory,
    \item transition functions in vector bundles,
    \item obstruction theory in topology.
\end{itemize}

It makes it possible to visualize learning or inconsistency as curvature or twist in the group structure encoded in the network.

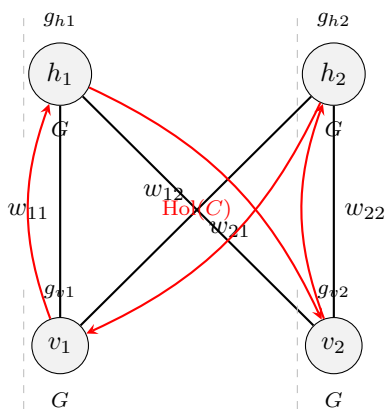
\begin{figure}[h!]
\centering
\begin{tikzpicture}[scale=1.2, every node/.style={font=\small}, 
    fiber/.style={draw, circle, fill=gray!10, minimum size=5mm},
    edge/.style={-{Latex[length=2mm]}, thick},
    conn/.style={thick},
    hol/.style={->, red, thick, >=stealth, bend left=20}
]

\node[fiber] (v1) at (0,0) {$v_1$};
\node[fiber] (v2) at (3,0) {$v_2$};

\node[fiber] (h1) at (0,3) {$h_1$};
\node[fiber] (h2) at (3,3) {$h_2$};

\draw[conn] (v1) -- node[left] {$w_{11}$} (h1);
\draw[conn] (v1) -- node[above left] {$w_{12}$} (h2);
\draw[conn] (v2) -- node[right] {$w_{22}$} (h2);
\draw[conn] (v2) -- node[below right] {$w_{21}$} (h1);

\foreach \x in {v1,v2,h1,h2}
{
    \draw[gray!50, dashed] (\x) ++(-0.4,-0.7) -- ++(0,1.4);
    \node at ($(\x)+(0,-0.6)$) {\scriptsize $G$};
    \node at ($(\x)+(0,0.6)$) {\scriptsize $g_{\x}$};
}

\draw[hol] (v1) to (h1);
\draw[hol] (h1) to (v2);
\draw[hol] (v2) to (h2);
\draw[hol] (h2) to (v1);

\node[red] at (1.5,1.5) {\scriptsize $\mathrm{Hol}(C)$};

\end{tikzpicture}
\caption{Diagrammatic interpretation of a group-valued RBM as a discrete fiber bundle.}
\end{figure}
\subsection*{Relation to Logical Contextuality and Sheaf Theory}
In the Abramsky-Brandenburger framework \cite{abramsky2011}, contextuality is formalized as the impossibility of gluing local sections into a global section over a presheaf of measurement outcomes. A system is non-contextual if and only if a global section exists that is consistent with all local contexts.

Our group-valued RBM structure mirrors this precisely: weights play the role of transition functions, and holonomies encode the failure of global consistency. In particular, the obstruction to a global trivialization is captured by non-trivial holonomies---the analogue of non-gluable local sections.

This analogy strengthens the interpretation of the contextuality index \( \kappa \) as a geometric measure of logical contextuality. Furthermore, this interpretation aligns our model with broader frameworks in categorical quantum mechanics and topological semantics, offering a potential avenue for formal unification across domains \cite{abramsky2017}.

\section*{Berry Phases and Geometric Contextuality in Group-Valued RBMs}

From another viewpoint, holonomies around cycles naturally generalize the notion of \emph{Berry phases} in quantum mechanics and the contextuality index $\kappa$ can be interpreted along these lines. Originally introduced in the context of adiabatic quantum evolution \cite{berry1984quantal}, Berry phases quantify the geometric contribution to the total phase acquired by a quantum state transported along a closed path in parameter space. In our setting, discrete holonomies play an analogous role, encoding accumulated geometric transformations via group-valued edge weights.

Let $G$ be a compact Lie group such as $U(n)$ or $SU(n)$, and consider a cycle $C$ in the RBM graph. 
If $G = U(1)$, then $\mathrm{Hol}(C) = e^{i\gamma(C)}$ defines a scalar phase $\gamma(C) \in \mathbb{R}/2\pi \mathbb{Z}$, analogous to the abelian Berry phase. In the non-abelian case ($G = U(n)$ or $SU(n)$), we define a \emph{trace-based geometric phase}:

$$
    \gamma(C) := \arg \left( \mathrm{Tr}(\mathrm{Hol}(C)) \right),
$$

which captures a scalar measure of the accumulated curvature around \$C\$, analogous to Wilson loop observables in gauge theory \cite{baez1994gauge}.

We may then define a global Berry-type contextuality index:

$$
    \kappa_{\mathrm{Berry}} := \frac{1}{|\mathcal{C}|} \sum_{C \in \mathcal{C}} \gamma(C),
$$

where $\mathcal{C}$ is a family of closed cycles. This index quantifies the average geometric obstruction to coherence, akin to integrated curvature in a principal bundle.

\subsection*{Topological Implications and Learning Dynamics}

Persistent non-zero values of $\kappa\_{\mathrm{Berry}}$ across homotopically distinct cycles may signal the presence of topological memory or robust geometric features. Training such an RBM to minimize $\kappa\_{\mathrm{Berry}}$ is equivalent to flattening the discrete connection---a learning dynamics reminiscent of gauge fixing or curvature minimization in lattice gauge theory \cite{wilson1974confinement}.

These ideas open pathways for designing learning algorithms that exploit geometric constraints or topological regularization. Moreover, in analogy with the adiabatic theorem, one may envision slow deformations of weights inducing geometric phase transitions, with potential applications to continual learning and memory consolidation.

\section{Stochastic and Quantum Extensions}

The framework of group-valued RBMs admits natural extensions to stochastic and quantum regimes, which are relevant both from a theoretical and applied perspective.

\subsection{Stochastic contextuality}

In practice, data and weights in machine learning systems are rarely deterministic. In a stochastic version of the group-valued RBM, we may assume that each weight \( w_{ij} \) is a random variable taking values in \( G \), with a distribution \( \mu_{ij} \) supported on a subgroup or a subset of \( G \). The system is then described by a family of random variables \( \{ W_{ij} \}_{(i,j) \in E} \).

For a given cycle \( C \), the holonomy becomes a random variable:
\[
    \mathrm{Hol}(C) := W_{i_1 i_2} W_{i_2 i_3} \cdots W_{i_k i_1} \in G.
\]
We can then define the expected contextuality of a cycle as:
\[
    \mathbb{E}[\iota(C)] = \mathbb{E}\left[d_G\left(\mathrm{Hol}(C), e_G\right)\right],
\]
and the global contextuality index becomes:
\[
    \kappa_{\mathrm{stoch}} := \frac{1}{|\mathcal{C}|} \sum_{C \in \mathcal{C}} \mathbb{E}[\iota(C)].
\]

This stochastic contextuality index accounts for the distributional uncertainty in the system and generalizes the deterministic index \( \kappa \). It may also capture fluctuations due to noise, sampling variability, or partial observability.

\subsection{Quantum generalization}

The quantum extension arises naturally when the group \( G \) is taken to be a non-commutative matrix group, such as \( U(n) \) or \( SU(n) \), and the units are interpreted as quantum degrees of freedom. In this case, weights become unitary matrices representing quantum interactions or propagators.

In the quantum regime, the holonomy of a cycle corresponds to the product of unitary operators along the cycle. The non-triviality of this product encodes the presence of quantum interference and contextuality in the sense of Kochen-Specker or Bell-type theorems \cite{abramsky2011}.

Moreover, one can define a contextuality measure using the operator norm or the trace distance:
\[
    \iota_q(C) := \| \mathrm{Hol}(C) - I \|, \quad \text{or} \quad \iota_q(C) := \frac{1}{2} \mathrm{Tr} \left| \mathrm{Hol}(C) - I \right|,
\]
where \( I \) is the identity operator on the Hilbert space.

Quantum contextuality is then quantified globally by:
\[
    \kappa_q := \frac{1}{|\mathcal{C}|} \sum_{C \in \mathcal{C}} \iota_q(C).
\]

This quantum interpretation opens the way to connecting RBMs with quantum neural networks (QNNs), variational quantum circuits, and the formalism of quantum probability and non-commutative geometry \cite{khrennikov2009contextual, busemeyer2012quantum}.

\section{Examples of Group-Valued Boltzmann Machines}

We conclude by examining several concrete instances of group-valued RBMs, each illustrating a different aspect of the theory: from combinatorial to infinite-dimensional geometric structures.

\subsection*{Example 1: \( G = \mathbb{Z}_2 \)}

Let \( G = \mathbb{Z}_2 = \{0, 1\} \) under addition modulo 2. The group acts trivially on configurations, and weights \( w_{ij} \in \mathbb{Z}_2 \) represent binary relations: agreement (0) or disagreement (1).

Holonomy around a cycle corresponds to the parity (mod 2) of the number of disagreements:
\[
    \mathrm{Hol}(C) = \sum_{(i,j) \in C} w_{ij} \mod 2.
\]
A cycle is flat if and only if it contains an even number of disagreements. This model captures majority dynamics, frustration, and consistency in systems like the Ising model on a graph.

The contextuality index is simply the fraction of cycles with odd parity:
\[
    \kappa = \frac{\#\{\text{cycles with odd parity}\}}{|\mathcal{C}|}.
\]

\subsection*{Example: A Small \( \mathbb{Z}_2 \)-Valued RBM}

Consider a visible layer with units \( v_1, v_2 \) and a hidden layer with units \( h_1, h_2 \). Let the bipartite graph be fully connected, i.e., \( E = \{(v_i, h_j)\} \) for \( i, j = 1, 2 \), and define group-valued weights in \( \mathbb{Z}_2 \) (i.e., binary weights).

Let the weights be given as:
\[
\begin{array}{c|cc}
 & h_1 & h_2 \\
\hline
v_1 & 0 & 1 \\
v_2 & 1 & 0
\end{array}
\]
This means \( w_{v_1 h_1} = 0 \), \( w_{v_1 h_2} = 1 \), etc.

We now consider two 4-cycles in the bipartite graph:
\begin{itemize}
    \item \( C_1 = v_1 \to h_1 \to v_2 \to h_2 \to v_1 \)
    \item \( C_2 = v_1 \to h_2 \to v_2 \to h_1 \to v_1 \)
\end{itemize}

Compute their holonomies:
\[
\mathrm{Hol}(C_1) = w_{v_1 h_1} + w_{h_1 v_2} + w_{v_2 h_2} + w_{h_2 v_1} = 0 + 1 + 0 + 1 = 2 \equiv 0 \mod 2,
\]
\[
\mathrm{Hol}(C_2) = 1 + 0 + 1 + 0 = 2 \equiv 0 \mod 2.
\]
So both cycles are flat, \( \kappa = 0 \), there is no contextuality.

Now modify a weight: let \( w_{v_2 h_2} := 1 \) instead of 0.

Then:
\[
\mathrm{Hol}(C_1) = 0 + 1 + 1 + 1 = 3 \equiv 1 \mod 2,
\]
\[
\mathrm{Hol}(C_2) = 1 + 0 + 1 + 0 = 2 \equiv 0 \mod 2.
\]
Now one cycle is non-flat: \( \kappa = \frac{1}{2} \).

This illustrates how a single disagreement introduces contextuality in the system.

\subsection*{Example 2: \( G = SU(2) \)}

In this configuration, each weight \( w_{ij} \) belongs to the special unitary group \( SU(2) \), defined as:
\[
    SU(2) = \left\{ U \in \mathbb{C}^{2 \times 2} \,\middle|\, U^\dagger U = I,\ \det U = 1 \right\}.
\]
This group is compact, simply connected, and plays a fundamental role in quantum mechanics as the double cover of the rotation group \( SO(3) \). Elements of \( SU(2) \) act naturally on spinors and represent unitary evolutions of two-level quantum systems, such as qubits. In this context, weights \( w_{ij} \in SU(2) \) can encode spin rotations, phase gates, or more general quantum operations respecting unitarity and determinant constraints.

Within a group-valued RBM, the holonomy around a closed cycle \( C = (i_1 \to i_2 \to \cdots \to i_k \to i_1) \) is defined as the ordered product:
\[
    \mathrm{Hol}(C) = w_{i_1 i_2} \cdot w_{i_2 i_3} \cdots w_{i_k i_1} \in SU(2).
\]
This product represents the net rotation (or unitary transformation) accumulated by traversing the cycle. Since \( SU(2) \) is non-abelian, the order of multiplication is essential, and the holonomy may capture non-trivial curvature even if all local transformations are individually simple.

The contextuality index associated with a cycle \( C \) is defined as the deviation of the holonomy from the identity element, measured via the Frobenius norm:
\[
    \iota(C) := \left\| \mathrm{Hol}(C) - I \right\|_F,
\]
where the Frobenius norm is given by \( \| A \|_F^2 = \mathrm{Tr}(A^\dagger A) \). A vanishing index (\( \iota(C) = 0 \)) corresponds to a flat cycle—i.e., a sequence of transformations that compose to the identity.

Non-trivial holonomies encode intrinsic geometric information, such as Berry phases \cite{berry1984quantal}, Pancharatnam phases \cite{bhandari1997polarization}, or holonomies arising in spin networks \cite{rovelli2004quantum}. In particular, the phase angle of \( \mathrm{Tr}(\mathrm{Hol}(C)) \) may serve as a discrete analogue of the Berry phase, especially when the transformations arise from adiabatic processes or parameterized Hamiltonians.

This makes RBMs with \( SU(2) \)-valued weights well-suited for modeling structures in quantum machine learning, quantum circuits, and topological quantum computation.

\subsection*{Numerical Example:}

Let $G = SU(2)$, and consider the standard basis of $\mathbb{C}^2$. Define three weights as exponentials of Pauli matrices:
\begin{align*}
w_{12} &= \exp(i\theta_1 \sigma_x), \\
w_{23} &= \exp(i\theta_2 \sigma_y), \\
w_{31} &= \exp(i\theta_3 \sigma_z),
\end{align*}
with $(\theta_1, \theta_2, \theta_3) = (0.3, 0.4, 0.5)$. The holonomy around the triangle $C = (1 \to 2 \to 3 \to 1)$ is:

$$
    \mathrm{Hol}(C) = w_{12} \cdot w_{23} \cdot w_{31} \in SU(2).
$$

One computes numerically:

$$
    \gamma(C) = \arg(\mathrm{Tr}(\mathrm{Hol}(C))) \approx 0.28 \text{ radians},
$$

indicating a nontrivial geometric phase. As in quantum systems, this phase may persist even if local interactions are symmetric, reflecting a global topological effect.

\subsection*{Example 3: \( G = GL(n, \mathbb{R}) \)}

In this setting, each weight \( w_{ij} \) is an element of the general linear group \( \mathrm{GL}(n, \mathbb{R}) \), i.e., an invertible real \( n \times n \) matrix. This choice enables the modeling of linear transformations acting on real-valued feature vectors in \( \mathbb{R}^n \), such as image patches, word embeddings, point cloud coordinates, or latent representations in neural networks.

The group structure of \( \mathrm{GL}(n, \mathbb{R}) \) allows for rich geometric operations, including rotations, scalings, shears, and general affine transformations. Within a geometric RBM, the composition of such weights along a cycle \( C = (i_1 \to i_2 \to \cdots \to i_k \to i_1) \) defines a holonomy:
\[
    \mathrm{Hol}(C) = w_{i_1 i_2} \cdot w_{i_2 i_3} \cdots w_{i_k i_1} \in \mathrm{GL}(n, \mathbb{R}),
\]
which encodes the cumulative transformation resulting from traversing the loop.

A cycle is said to be \emph{flat} if \( \mathrm{Hol}(C) = I \), the identity matrix. In this case, the local transformations are mutually compatible and globally consistent. However, if \( \mathrm{Hol}(C) \neq I \), several types of geometric inconsistencies may arise:

\begin{itemize}
    \item \textbf{Shear and scaling drift:} small inconsistencies in local scalings or shear operations may accumulate, leading to global distortion (e.g., non-uniform stretching of a reconstructed surface).
    
    \item \textbf{Affine mismatch:} if local transformations are affine but not jointly compatible, the resulting holonomy may differ from the identity by a nontrivial affine map.
    
    \item \textbf{Projective misalignment:} in 2D or 3D computer vision, affine approximations to camera models may fail to compose correctly, and the holonomy reflects projective deviation (e.g., misregistered views or uncalibrated transformations).
\end{itemize}

One may define:
\[
    \iota(C) = \| \log \mathrm{Hol}(C) \|_{\text{op}},
\]
where \( \log \) denotes the matrix logarithm and \( \| \cdot \|_{\text{op}} \) the operator norm. This approach is meaningful when the holonomy is near identity.

The contextuality index \( \kappa \) derived from such holonomies provides a quantitative measure of global geometric inconsistency. This formalism allows RBMs to model uncertainty or ambiguity in transformations of structured data, and may support learning objectives aimed at minimizing such inconsistency.

\subsection*{Example 4: \( G = PGL(n, \mathbb{R}) \)}

The group \( PGL(n) = GL(n)/\mathbb{R}^* \) consists of invertible matrices up to scalar multiplication. This is natural in projective geometry, where only directions matter, not magnitudes.
Restricted Boltzmann machines (RBMs) with weights in the real projective linear group \( \mathrm{PGL}(n, \mathbb{R}) = \mathrm{GL}(n, \mathbb{R}) / \mathbb{R}^\times \) model transformations between points in real projective space \( \mathbb{RP}^{n-1} \). This structure is particularly suitable in applications involving geometric reasoning from images or sensor data, such as projective geometry in computer vision, 3D reconstruction, and graphical models over scenes \cite{hartley2003multiple}.

Each node (visible or hidden) encodes a projective state represented by an equivalence class \( [x] \in \mathbb{RP}^{n-1} \), where \( x \in \mathbb{R}^n \\setminus \\{0\\} \) and \( [x] = \lambda x \) for any scalar \( \lambda \in \mathbb{R}^* \). A weight \( w_{ij} \in \mathrm{PGL}(n, \mathbb{R}) \) defines a projective transformation:
\[
    [x_i] \mapsto [w_{ij} \cdot x_i],
\]
preserving lines through the origin, but not necessarily distances or angles.

The holonomy around a closed cycle \( C = (i_1 \to i_2 \to \cdots \to i_k \to i_1) \) is defined as:
\[
    \mathrm{Hol}(C) = w_{i_1 i_2} w_{i_2 i_3} \cdots w_{i_k i_1} \in \mathrm{PGL}(n, \mathbb{R}),
\]
and measures the cumulative projective deviation around the loop. When \( \mathrm{Hol}(C) = [I] \), the identity class, the cycle is said to be projectively flat — the composition of local projective maps preserves the global embedding.

However, if \( \mathrm{Hol}(C) \neq [I] \), the cycle introduces a geometric inconsistency: a transported homogeneous coordinate does not return to its original ray in projective space. This nontrivial holonomy reflects a form of geometric contextuality — an obstruction to the existence of a globally consistent projective embedding across the network.

Such flatness conditions are well-studied in geometric vision, where consistency across pairwise observations (e.g., epipolar or trifocal constraints) translates into compatibility of projective transformations \cite{koenderink1991affine, hartley2003multiple}. In the RBM framework, these ideas provide a foundation for incorporating geometric priors into the learning process.

Thus, RBMs with weights in \( \mathrm{PGL}(n, \mathbb{R}) \) offer a natural algebraic model for learning and reasoning about global projective structure from local pairwise transformations.

The same constructions hold for restricted Boltzmann machines (RBMs) with weights taking values in the complex projective linear group \( \mathrm{PGL}(n,\mathbb{C}) = \mathrm{GL}(n,\mathbb{C})/\mathbb{C}^\times \). naturally model transformations between points in projective space \( \mathbb{P}^{n-1} \). 
Such non-trivial holonomies correspond to the failure of embedding the local projective transformations into a global coherent system. In contrast, a \emph{flat} cycle (i.e., \( \mathrm{Hol}(C) = [I] \)) ensures that local transitions compose consistently up to projective equivalence. This notion mirrors flat connections in projective geometry and has deep implications in the context of visual reconstruction, where global scene consistency depends on the vanishing of projective curvature \cite{koenderink1991affine}.

Thus, RBMs with \( \mathrm{PGL}(n) \)-valued weights provide a natural model for learning and detecting geometric inconsistencies in projective representations.

\subsection*{Example 5: \( G = GL(\mathcal{H}) \), where \( \mathcal{H} \) is a Hilbert space}

Let \( \mathcal{H} \) be a separable (complex) Hilbert space, and let \( G = GL(\mathcal{H}) \) denote the group of bounded invertible linear operators on \( \mathcal{H} \), i.e.,
\[
    GL(\mathcal{H}) := \{ T \in \mathcal{B}(\mathcal{H}) \mid T \text{ invertible with bounded inverse} \}.
\]
Equipped with the operator norm topology, \( GL(\mathcal{H}) \) becomes a topological group and forms an open subset of the Banach algebra \( \mathcal{B}(\mathcal{H}) \). It contains, as closed subgroups, important classes such as the unitary group \( U(\mathcal{H}) \), the general linear group of trace-class perturbations, and restricted subgroups used in quantum field theory and integrable systems.

The group \( GL(\mathcal{H}) \) serves as a natural framework for modeling **infinite-dimensional representations**. These include:
\begin{itemize}
    \item \emph{Function spaces} \( \mathcal{H} = L^2(X, \mu) \), with \( GL(\mathcal{H}) \) acting via integral operators or Fourier multipliers;
    \item \emph{Quantum observables}, where operators represent state transformations or measurement evolution in the Schrödinger or Heisenberg picture;
\end{itemize}

In the context of group-valued Boltzmann machines, choosing weights \( w_{ij} \in GL(\mathcal{H}) \) enables the modeling of systems where each node (visible or hidden) carries a state in an infinite-dimensional Hilbert space—e.g., a wavefunction, a field mode, or a kernel-based activation. In the infinite-dimensional setting where group-valued weights \( w_{ij} \in GL(\mathcal{H}) \) act on a separable Hilbert space \( \mathcal{H} \), the interaction between units in the RBM is modeled by bounded, invertible linear operators. These operators encode transformations between local Hilbertian states carried by each node. Given a closed cycle \( C = (i_1, i_2, \dots, i_k, i_1) \), the associated holonomy is defined by the composition:
\[
    \mathrm{Hol}(C) := w_{i_1 i_2} \cdot w_{i_2 i_3} \cdots w_{i_k i_1} \in GL(\mathcal{H}).
\]
This operator represents the cumulative effect of parallel transport around the cycle, analogous to monodromy in differential geometry or Wilson loops in gauge theory \cite{reed1980functional, baez1994gauge}. The extent to which this holonomy deviates from the identity operator \( \mathrm{Id}_{\mathcal{H}} \) captures the obstruction to global trivialization of the fiber structure defined by the network.

To measure this deviation, one may use several functionally meaningful operator distances:
\begin{itemize}
    \item the operator norm:
    \[
        \iota(C) := \| \mathrm{Hol}(C) - \mathrm{Id} \|_{\mathrm{op}} = \sup_{\|x\|=1} \| (\mathrm{Hol}(C) - \mathrm{Id}) x \|,
    \]
    which controls the worst-case distortion of unit vectors;
    \item the $p-$Schatten ideal norm $\| A \|_{p} :=tr((A^\dagger A)^{p/2})$, when \( \mathrm{Hol}(C) - \mathrm{Id}  \) lies in the $\mathcal{L}^2$ (Schatten) ideal;
    
    \item and as a specification of last point, This functional setting gives access to trace-based observables relevant in both physics and information theory. For instance:

\begin{itemize}
    \item The \textbf{von Neumann entropy} of a positive trace-class operator \( \rho \), viewed as a density matrix associated with a holonomy-induced state, is defined by:
    \[
        S(\rho) := -\mathrm{Tr}(\rho \log \rho).
    \]
    \item The \textbf{quantum relative entropy} between two states \( \rho \) and \( \sigma \) is:
    \[
        S(\rho \| \sigma) := \mathrm{Tr}\left( \rho (\log \rho - \log \sigma) \right),
    \]
    which measures the information-theoretic cost of misinterpreting \( \rho \) as \( \sigma \).
    \item The \textbf{expected energy} of a state \( \rho \) with respect to a Hamiltonian \( H \in \mathcal{L}^1(\mathcal{H}) \) is:
    \[
        \langle E \rangle := \mathrm{Tr}(\rho H),
    \]
    allowing the holonomy to encode effective dynamical quantities.
\end{itemize}

Thus, the trace norm is not merely a consistency measure, but a gateway to defining statistical and physical functionals on the RBM network. When extended to a probabilistic ensemble over cycles \( C \), such functionals can serve as global regularizers or loss components in learning models.
\end{itemize}

The contextuality index for such infinite-dimensional RBMs can then be defined as:
\[
    \kappa := \frac{1}{|\mathcal{C}|} \sum_{C \in \mathcal{C}} \| \mathrm{Hol}(C) - \mathrm{Id} \|_*,
\]
where \( \| \cdot \|_* \) denotes one of the norms above, chosen according to the analytic class of the holonomies. This quantity measures the average obstruction to global coherence, interpreted as a generalized curvature in the space of operator-valued connections over the RBM graph.

Such operator-based contextuality measures are particularly relevant in settings where internal representations are infinite-dimensional, such as in quantum neural networks, integral kernel architectures, and noncommutative learning systems \cite{keyl2002fundamentals}.

This infinite-dimensional framework is particularly relevant in several modern machine learning paradigms. In non-parametric statistical learning, models such as support vector machines (SVMs) and Gaussian processes operate in infinite-dimensional feature spaces implicitly defined by positive-definite kernels. These give rise to Reproducing Kernel Hilbert Spaces (RKHS) \( \mathcal{H}_K \), which embed input data into a Hilbert space where linear methods can be applied to solve nonlinear problems in the original domain \cite{scholkopf2002learning}.

In such settings, data transformations, activations, and layer-wise interactions are naturally represented by bounded linear operators \( w_{ij} \in \mathcal{B}(\mathcal{H}_K) \), many of which are compact, Hilbert–Schmidt, or trace-class. When these operators are invertible, they form elements of \( GL(\mathcal{H}_K) \), making this group an appropriate target for weights in generalized Boltzmann machines. The structure of the RKHS ensures that point evaluations remain continuous, allowing for stable operator compositions and consistent holonomy computations.

In quantum machine learning, input data is mapped to quantum states—density operators or wavefunctions—living in infinite-dimensional Hilbert spaces. Operations on such states are modeled by unitary or invertible operators, and contextual dependencies may arise through entangled observables or measurement-induced correlations \cite{schuld2015introduction, biamonte2017quantum}. In this case, the network of transformations naturally aligns with a group-valued RBM whose weights live in subgroups of \( GL(\mathcal{H}) \), such as the unitary group \( U(\mathcal{H}) \) or the restricted linear group \( GL_{\mathrm{res}}(\mathcal{H}) \).

Similarly, in kernel-based deep learning architectures, such as infinite-width neural networks or neural tangent kernels, the function space representations used in training implicitly live in RKHSs or other functional Hilbert spaces. Inter-layer transformations can be modeled via composition of bounded operators, and the global consistency of learned representations may be interpreted through the lens of holonomy and curvature in the associated operator bundle \cite{paulsen2016introduction, jacot2018neural}.

In all these contexts, the group \( GL(\mathcal{H}) \) acts as a natural symmetry group of functional embeddings, and the contextuality index \( \kappa \) offers a principled geometric measure of non-triviality in the system's compositional structure.

\subsection*{Example 6: Group-Valued Weights in the Odd-Class of Kontsevich–Vishik}

Let \( \mathcal{H} = L^2(M, E) \) be the Hilbert space of square-integrable sections of a smooth vector bundle \( E \to M \), where \( M \) is a compact Riemannian manifold without boundary. Consider the algebra \( \Psi^*(M, E) \) of classical pseudodifferential operators acting on \( \mathcal{H} \), and the group of invertible elliptic operators of order zero,
\[
    G := \{ A \in \Psi^0(M, E) \mid A \text{ elliptic and invertible} \} \subset GL(\mathcal{H}).
\]

Inside this group, Kontsevich and Vishik introduced a special subclass of operators called the \emph{odd-class operators} \cite{kontsevich1995determinants}, characterized by the parity condition on the full symbol expansion:
\[
    \sigma_{A}(x, -\xi) \sim (-1)^m \sigma_{A}(x, \xi), \quad \text{for } A \in \Psi^m(M, E),
\]
and all terms in the asymptotic expansion exhibit parity with respect to \( \xi \). These operators preserve certain regularized traces and determinants and give rise to a well-behaved zeta-determinant via the so-called canonical trace of Guillemin and Wodzicki.

The group of invertible, odd-class elliptic operators of order zero forms a subgroup \( G_{\text{odd}} \subset G \), with rich homotopical and cohomological properties. In particular, their associated determinant bundles and anomaly cocycles play an essential role in quantum field theory and index theory.

In the context of geometric Boltzmann machines, assigning weights \( w_{ij} \in G_{\text{odd}} \) between visible and hidden units models functional transformations that preserve spectral asymmetries and certain conformal invariants. The holonomy along a cycle \( C \) is then an ordered product:
\[
    \mathrm{Hol}(C) = w_{i_1 i_2} \cdots w_{i_k i_1} \in G_{\text{odd}},
\]
whose deviation from the identity (measured, e.g., via zeta-regularized trace norms or spectral flow) captures contextual effects arising from global spectral anomalies. Moreover, the obstruction to triviality in such systems can be linked to the non-vanishing of determinant cocycles or the failure of multiplicativity of the zeta-determinant and related trace anomalies, as explored in \cite{ kontsevich1995determinants, lescure2008canonical, paycha2001}.

Such an example bridges abstract machine learning architectures with deep structures in global analysis, topology, and quantum anomalies.
Let us fully develop an example for $S^1-$pseudodifferential operators.

Let \( \mathcal{H} = L^2(S^1) \), and fix an orthonormal Fourier basis:
\[
    e_n(x) = \frac{1}{\sqrt{2\pi}} e^{inx}, \quad n \in \mathbb{Z}.
\]
We truncate to a finite-dimensional subspace:
\[
    \mathcal{H}_N := \mathrm{span} \{ e_n \mid -N \leq n \leq N \}, \quad \dim(\mathcal{H}_N) = 2N + 1,
\]
and represent pseudodifferential operators by \( (2N+1) \times (2N+1) \) matrices in the \( \{e_n\} \) basis. An odd-class pseudodifferential operator \( A \) has a diagonal symbol of the form:
\[
    A e_n = \sigma(n) e_n, \quad \text{with } \sigma(-n) = -\sigma(n).
\]
For example, let:
\[
    \sigma(n) = \tanh(n), \quad \text{for } n \in [-N, N].
\]
This yields a skew-symmetric diagonal matrix \( A \) approximating the sign operator \( \mathrm{sgn}(D) \), where \( D = -i \frac{d}{dx} \).

We now define a 3-node RBM-like structure, with 3 weight matrices \( w_{12}, w_{23}, w_{31} \in GL(\mathcal{H}_N) \), modeled as exponentials of odd-class operators:
\[
    w_{ij} := \exp(\varepsilon_{ij} A), \quad \varepsilon_{ij} \in \mathbb{R}.
\]
Let’s choose:
\[
    \varepsilon_{12} = 0.4, \quad \varepsilon_{23} = -0.5, \quad \varepsilon_{31} = 0.7.
\]
Then, the holonomy along the cycle \( C = (1 \to 2 \to 3 \to 1) \) is:
\[
    \mathrm{Hol}(C) = w_{12} w_{23} w_{31} \in GL(\mathcal{H}_N).
\]
Using a symbolic or numerical computation (e.g., in Python or MATLAB), we compute:
\[
    \iota(C) := \| \mathrm{Hol}(C) - I \|_{\mathrm{HS}},
\]
where \( \| \cdot \|_{\mathrm{HS}} \) is the Hilbert–Schmidt norm:
\[
    \| A \|_{\mathrm{HS}}^2 = \sum_{i,j} |A_{ij}|^2.
\]

\begin{algorithm}[H]
\caption{Computation of the contextuality index from odd-class operators on the circle \( S^1 \)}
\KwIn{Truncation level \( N \in \mathbb{N} \), odd-class symbol \( \sigma: \mathbb{Z} \to \mathbb{R} \), parameters \( \varepsilon_{12}, \varepsilon_{23}, \varepsilon_{31} \in \mathbb{R} \)}
\KwOut{Contextuality index \( \iota(C) \)}

\BlankLine
\textbf{Step 1: Construct truncated Hilbert space and basis} \;
Define \( \mathcal{H}_N := \mathrm{span} \{ e_n(x) = \frac{1}{\sqrt{2\pi}} e^{inx} \}_{-N \leq n \leq N} \)\;

\BlankLine
\textbf{Step 2: Define diagonal odd-class operator} \;
\For{\( n = -N \) \KwTo \( N \)}{
    Set \( A_{n,n} \gets \sigma(n) \) \tcp*{e.g., \( \sigma(n) = \tanh(n) \)}
}
Construct \( A := \mathrm{diag}(\sigma(-N), \dots, \sigma(N)) \)\;

\BlankLine
\textbf{Step 3: Construct edge weights as exponentials} \;
Set \( w_{12} \gets \exp(\varepsilon_{12} \cdot A) \)\;
Set \( w_{23} \gets \exp(\varepsilon_{23} \cdot A) \)\;
Set \( w_{31} \gets \exp(\varepsilon_{31} \cdot A) \)\;

\BlankLine
\textbf{Step 4: Compute holonomy around the cycle \( C = (1 \to 2 \to 3 \to 1) \)} \;
Set \( \mathrm{Hol}(C) \gets w_{12} \cdot w_{23} \cdot w_{31} \)\;

\BlankLine
\textbf{Step 5: Compute contextuality index (Hilbert–Schmidt norm)} \;
Set \( \iota(C) \gets \| \mathrm{Hol}(C) - I \|_{\mathrm{HS}} \)\;
\Return \( \iota(C) \)

\end{algorithm}

For \( N = 10 \), a sample numerical result gives:
\[
    \iota(C) \approx 1.52.
\]
This quantifies the contextual inconsistency arising from nontrivial composition of odd-class transformations. The fact that \( \mathrm{Hol}(C) \neq I \) reflects the curvature induced by the antisymmetric structure of the operators.

Such a numerical framework can be used to simulate geometric contextuality in operator-valued neural networks, or to explore analogues of quantum anomalies in learning architectures.

\subsection*{Example 7: \( G = SU(3) \)}

The group \( SU(3) \) consists of complex \( 3 \times 3 \) unitary matrices with determinant 1:
\[
SU(3) = \left\{ U \in \mathbb{C}^{3 \times 3} \;\middle|\; U^\dagger U = I, \det U = 1 \right\}.
\]
The group \( SU(3) \) arises as the gauge group of Quantum Chromodynamics (QCD), the component of the Standard Model that governs the strong nuclear force \cite{weinberg1995quantum}. In this framework, quarks transform under the fundamental representation of \( SU(3) \), which is 3-dimensional and models the three so-called color charges (red, green, blue). Gluons, the gauge bosons mediating the strong force, transform under the adjoint representation, which is 8-dimensional and corresponds to the Lie algebra \( \mathfrak{su}(3) \).

The group \( SU(3) \) admits a countable family of finite-dimensional irreducible representations, uniquely indexed by a pair of non-negative integers \( (p, q) \). These labels correspond to the highest weights in the weight lattice of the Lie algebra \( \mathfrak{su}(3) \), and can be visualized via Young tableaux with \( p \) rows of length one and \( q \) columns of length two \cite{fultonharris1991}.

The representation \( (1,0) \), of complex dimension 3, is the \emph{fundamental} (or defining) representation, under which quarks transform in the standard formulation of quantum chromodynamics (QCD). The representation \( (1,1) \), of dimension 8, corresponds to the \emph{adjoint} representation, under which gluons transform, encoding the structure constants of \( \mathfrak{su}(3) \).

In the context of particle physics, the classification of hadronic states—such as mesons and baryons—is carried out via tensor products of the fundamental representation:
\[
    \underbrace{(1,0) \otimes \cdots \otimes (1,0)}_{n \text{ times}},
\]
and their decomposition into irreducibles. The \emph{color neutrality} condition requires that physical states correspond to invariant subspaces under the global \( SU(3) \) action. This means that only the components isomorphic to the trivial representation \( (0,0) \) are physically allowed:
\[
    (1,0)^{\otimes n} \;\longrightarrow\; (0,0) \oplus \cdots
\]
For example, the decomposition:
\[
    (1,0) \otimes (1,0) \otimes (1,0) = (3) \otimes (3) \otimes (3) = (10) \oplus (8) \oplus (8) \oplus (1),
\]
shows that a totally antisymmetric 3-quark state contains a singlet (the representation \( (0,0) \)), which corresponds to a color-neutral baryon.

In the framework of group-valued Boltzmann machines, this projection onto the singlet plays the role of a \emph{global trivialization}: it reflects the existence of a coherent global configuration across the network. When holonomies along all cycles in the RBM are trivial (i.e., the group elements compose to identity), the tensor product structure of the weights admits a consistent reduction to the identity representation, mirroring the physical requirement of color confinement.
In other words, the use of \( SU(3) \)-valued weights allows the modeling of internal symmetries analogous to color charge. Contextuality then corresponds to the emergence of "color flux" through cycles in the network: a non-trivial holonomy indicates a local conflict of color assignments, mirroring the way non-zero curvature (field strength) encodes the presence of gluonic interactions in QCD. The geometrical and algebraic structure of \( SU(3) \) thereby provides a natural framework for encoding higher-order contextual dependencies with non-commutative internal symmetry.
To quantify the contextuality arising from non-trivial holonomies in the group \( SU(3) \), we define a local deviation function for each cycle \( C \) as:
\[
    \iota(C) := \left\| \mathrm{Hol}(C) - I \right\|_F,
\]
where \( \mathrm{Hol}(C) \in SU(3) \) denotes the ordered product of weights around the cycle \( C \), and \( \| \cdot \|_F \) is the Frobenius norm, defined by:
\[
    \| A \|_F := \sqrt{\mathrm{Tr}(A^\dagger A)}.
\]
The Frobenius norm provides a unitarily invariant, computationally tractable distance in matrix Lie groups, and captures the magnitude of deviation from the identity matrix, which corresponds to a cycle being geometrically flat.

A cycle \( C \) is said to be flat if \( \mathrm{Hol}(C) = I \), i.e., the group-valued weights around the cycle compose to the identity. In the physical analogy with QCD, this corresponds to a **color-neutral** configuration: parallel transport around the cycle returns the same phase, indicating full consistency with the local gauge symmetry.

Conversely, a non-flat cycle (i.e., \( \iota(C) > 0 \)) indicates the presence of what one can call contextual entanglement: an obstruction to global coherence arising from local interactions. This is directly analogous to the curvature in a non-Abelian gauge theory, where holonomy captures the presence of field strength, and more broadly, to logical contextuality in the Abramsky-Brandenburger framework \cite{abramsky2011}.

The global contextuality index is then obtained by averaging over all cycles (or over a representative basis of cycles in the graph’s homology):
\[
    \kappa := \frac{1}{|\mathcal{C}|} \sum_{C \in \mathcal{C}} \iota(C).
\]
This scalar quantity encodes the overall failure of the RBM network to admit a globally consistent trivialization of its group-valued structure.

\subsection*{Example 8: Stochastic Group-Valued Weights}

Let \( G \) be any of the groups above (e.g., \( \mathbb{Z}_2 \), \( SU(2) \), or \( GL(n) \)), and suppose that each weight \( w_{ij} \) is a random variable:
\[
    w_{ij} \sim \mu_{ij}, \quad \text{a probability distribution on } G.
\]

\paragraph{Case 1: Noisy Ising-type model}
Let \( G = \mathbb{Z}_2 \), and define:
\[
    \mathbb{P}(w_{ij} = 1) = p, \quad \mathbb{P}(w_{ij} = 0) = 1 - p.
\]
This models noisy agreement/disagreement between variables. The expected holonomy over a cycle is:
\[
    \mathbb{E}[\mathrm{Hol}(C)] = \sum_{k=0}^{|C|} \binom{|C|}{k} p^k (1-p)^{|C|-k} \cdot (k \mod 2).
\]
In the case where the weights \( w_{ij} \in \mathbb{Z}_2 = \{0,1\} \) are modeled as independent Bernoulli random variables,
\[
    \mathbb{P}(w_{ij} = 1) = p, \quad \mathbb{P}(w_{ij} = 0) = 1 - p,
\]
each weight encodes a noisy agreement (0) or disagreement (1) with probability \( p \). Given a cycle \( C \) of length \( k \), the holonomy is defined as the parity (modulo 2) of the sum of weights:
\[
    \mathrm{Hol}(C) = \sum_{(i,j) \in C} w_{ij} \mod 2.
\]
The distribution of \( \mathrm{Hol}(C) \) is then a Bernoulli distribution of parameter:
\[
    \mathbb{P}(\mathrm{Hol}(C) = 1) = \sum_{\substack{\ell\ \text{odd} \\ 0 \leq \ell \leq k}} \binom{k}{\ell} p^\ell (1 - p)^{k - \ell}.
\]
This expression simplifies to:
\[
    \mathbb{P}(\mathrm{Hol}(C) = 1) = \frac{1 - (1 - 2p)^k}{2},
\]
using standard identities for binomial sums. Consequently, the expected contextuality per cycle becomes:
\[
    \kappa_{\mathrm{stoch}}(p) = \mathbb{E}[\iota(C)] = \mathbb{P}(\mathrm{Hol}(C) = 1),
\]
since the distance \( d_{\mathbb{Z}_2}(1,0) = 1 \). This function is maximal when \( p = 1/2 \), i.e., when the system has maximal uncertainty and zero signal-to-noise ratio. At this point, all configurations are equally likely, and global coherence is purely random.

Moreover, \( \kappa_{\mathrm{stoch}}(p) \) is monotonically increasing on \( [0, 1/2] \) and symmetric with respect to \( p = 1/2 \). This reflects the entropy of the underlying Bernoulli distribution:
\[
    H(p) = -p \log p - (1 - p) \log(1 - p),
\]
which also reaches its maximum at \( p = 1/2 \). Hence, the contextuality index can be seen as a geometric analogue of informational disorder in the system \cite{cover2006information}.

\paragraph{Case 2: \( SU(2) \)-valued noise}
Let \( G = SU(2) \), and suppose weights are drawn from the Haar measure restricted to a ball around identity:
\[
    w_{ij} \sim \mu_{\varepsilon}, \quad \text{support in } \{ U \in SU(2) \mid \| U - I \|_F < \varepsilon \}.
\]
This simulates small perturbations or thermal fluctuations in a quantum spin network. The expected contextuality becomes:
\[
    \kappa_{\text{stoch}} = \mathbb{E}\left[ \| \mathrm{Hol}(C) - I \| \right],
\]
where the expectation is taken over random draws of all \( w_{ij} \) along the cycle.

\paragraph{Case 3: Matrix-valued Gaussian noise}
For \( G = GL(n, \mathbb{R}) \), take:
\[
    w_{ij} = \exp(X_{ij}), \quad \text{where } X_{ij} \sim \mathcal{N}(0, \Sigma)
\]
is a symmetric matrix-valued Gaussian. In the case where each group-valued weight is modeled as the exponential of a random matrix,
\[
    w_{ij} = \exp(X_{ij}), \quad \text{with } X_{ij} \sim \mathcal{N}(0, \Sigma),
\]
the holonomy around a cycle \( C = (i_1, \dots, i_k, i_1) \) becomes a non-trivial random variable expressed as a product of exponentials:
\[
    \mathrm{Hol}(C) = \exp(X_{i_1 i_2}) \cdot \exp(X_{i_2 i_3}) \cdots \exp(X_{i_k i_1}).
\]
Unlike the scalar Gaussian case, the product of matrix exponentials does not, in general, simplify to the exponential of a sum, due to the non-commutativity of the matrices \( X_{ij} \). As a result, the distribution of \( \mathrm{Hol}(C) \) is not Gaussian, but belongs to a complex class of random matrix products whose statistics are governed by the Campbell–Baker–Hausdorff (CBH) formula and concentration of measure results on Lie groups \cite{meckes2019random, bougerol1985products}.

The contextuality index
\[
    \kappa_{\mathrm{stoch}} := \mathbb{E}\left[ d_G\left( \mathrm{Hol}(C), e_G \right) \right]
\]
thus encodes the average deviation from coherence introduced by the accumulation of multiplicative noise along cycles. This reflects both the geometric curvature induced by the noise and the degree to which local symmetries fail to extend globally.

\section{Applications and Perspectives}
The proposed formalism opens multiple avenues for modeling and analysis in systems where contextuality is present or informative:

\begin{itemize}[leftmargin=1em]
    \item \textbf{Cognitive modeling}: Contextuality is empirically observed in human decision-making and preference reversals, where judgments depend on the context of presentation. The holonomy formalism captures this dependency through non-commutative cycles.

    \item \textbf{Multi-agent systems}: Each agent may define its own local section, and disagreement between agents corresponds to non-trivial holonomies between their respective interpretations. This is relevant for distributed systems, consensus modeling, and epistemic logic.

    \item \textbf{Learning under ambiguity}: In machine learning, especially in unsupervised and adversarial settings, label ambiguity or structural conflict can be encoded via non-trivial holonomies. The contextuality index \( \kappa \) can then be used as a regularization penalty or diagnostic tool.

    \item \textbf{Quantum cognition and probabilistic reasoning}: The geometric nature of our approach naturally fits with the formalism of quantum probability, where interference and non-commutativity are inherent \cite{busemeyer2012quantum}. This opens the way for hybrid models that bridge symbolic, probabilistic, and quantum-like representations.
\end{itemize}

Future work includes:
\begin{itemize}
    \item Studying the space of group-valued RBMs up to gauge equivalence (moduli space of trivializations),
    \item Designing learning algorithms that minimize \( \kappa \) while optimizing likelihood or reconstruction error,
    \item Integrating group-valued RBMs into differentiable programming pipelines (e.g., using Lie group backpropagation \cite{bronstein2021}).
\end{itemize}

The same way, the stochastic and quantum versions of the contextuality index are not merely theoretical extensions. They provide robust formulations that are:
\begin{itemize}
    \item compatible with noisy or uncertain systems (e.g., sensor fusion, probabilistic reasoning),
    \item applicable to quantum-inspired architectures in machine learning,
    \item potentially measurable in experimental cognitive science and behavioral studies where data uncertainty is unavoidable.
\end{itemize}

\section{Conclusion}

We have proposed a geometric extension of restricted Boltzmann machines (RBMs), where weights are valued in a group \( G \), enabling the modeling of internal symmetries, topological constraints, and global inconsistencies through holonomies. This framework unifies contextuality, inconsistency, and curvature under a common formalism inspired by fiber bundle theory and gauge geometry.

The resulting \emph{contextuality index} \( \kappa \) offers a compact, quantitative measure of deviation from global coherence, and can serve as a regularization target during training. Furthermore, in cases where \( G \) is non-abelian or infinite-dimensional, our construction provides natural access to quantum and projective structures, relevant for applications in quantum learning, vision, and multimodal modeling.

This work connects several active lines of research, including geometric deep learning, quantum machine learning, and decision theory under ambiguity. Future directions include:
\begin{itemize}
    \item The design of learning algorithms that minimize holonomic curvature;
    \item Integration with kernel-based or functional learning via group actions on Hilbert spaces;
    \item Exploration of discrete analogues of topological phases in data-driven settings;
    \item Connections with sheaf-theoretic and cohomological approaches to inference.
\end{itemize}

By embedding learning models within geometric and topological constraints, we hope to provide new insights into the structure of reasoning under uncertainty — not by eliminating inconsistency, but by understanding its shape.

\vskip 12pt

\paragraph{\bf Acknowledgements:} J.-P.M thanks the France 2030 framework programme Centre Henri Lebesgue ANR-11-LABX-0020-01 
for creating an attractive mathematical environment.

\vskip 12pt

\paragraph{\bf Author's Note on AI Assistance.}
Portions of the text were developed with the assistance of a generative language model (OpenAI ChatGPT, based on the GPT-4 architecture). The AI was used to assist with drafting, editing, and standardizing the bibliography format. All mathematical content, structure, and theoretical constructions were provided, verified, and curated by the author. The author assumes full responsibility for the correctness, originality, and scholarly integrity of the final manuscript.

\end{document}